\documentclass[letterpaper, 10 pt, conference]{ieeeconf}  % Comment this line out if you need a4paper

\IEEEoverridecommandlockouts                              % This command is only needed if 
\usepackage{graphics} % for pdf, bitmapped graphics files
\usepackage{amsmath} % assumes amsmath package installed
\usepackage{amssymb}  % assumes amsmath package installed
\usepackage{adjustbox}
\usepackage{array}
\usepackage{cite}
\usepackage{multirow}
\usepackage{makecell}
\usepackage{hyperref}
\usepackage{commath}
\hypersetup{
    colorlinks=true,
    linkcolor=magenta,
    filecolor=magenta,
    urlcolor=magenta,
}

\title{\LARGE \bf
Learning Spatial Context with Graph Neural Network\\for Multi-Person Pose Grouping
}

\author{Jiahao Lin$^{1}$ and Gim Hee Lee$^{1}$% <-this % stops a space
\thanks{$^{1}$Department of Computer Science, National University of Singapore, {\tt\small jiahao@comp.nus.edu.sg, gimhee.lee@nus.edu.sg}}%
}

\begin{document}

\maketitle
\thispagestyle{empty}
\pagestyle{empty}

%%%%%%%%%%%%%%%%%%%%%%%%%%%%%%%%%%%%%%%%%%%%%%%%%%%%%%%%%%%%%%%%%%%%%%%%%%%%%%%%
\begin{abstract}

Bottom-up approaches for image-based multi-person pose estimation consist of two stages: (1) keypoint detection and (2) grouping of the detected keypoints to form person instances. Current grouping approaches rely on learned embedding from only visual features that completely ignore the spatial configuration of human poses. In this work, we formulate the grouping task as a graph partitioning problem, where we learn the affinity matrix with a Graph Neural Network (GNN). More specifically, we design a Geometry-aware Association GNN that utilizes spatial information of the keypoints and learns local affinity from the global context.
The learned geometry-based affinity is further fused with appearance-based affinity to achieve robust keypoint association. Spectral clustering is used to partition the graph for the formation of the pose instances. Experimental results on two benchmark datasets show that our proposed method outperforms existing appearance-only grouping frameworks, which shows the effectiveness of utilizing spatial context for robust grouping.
Source code is available at: \url{https://github.com/jiahaoLjh/PoseGrouping}.

\end{abstract}

%%%%%%%%%%%%%%%%%%%%%%%%%%%%%%%%%%%%%%%%%%%%%%%%%%%%%%%%%%%%%%%%%%%%%%%%%%%%%%%%
\section{INTRODUCTION}

The problem of human pose estimation is of great interest in the field of computer vision.
Although it is a challenging problem, human pose estimation has many potential applications in augmented/virtual reality, human robot interaction, security surveillance, \textit{etc.}
With the rapid development of deep learning techniques such as Convolutional Neural Networks (CNNs), vision systems are getting increasingly capable of extracting high-level visual features for detecting keypoints of interest.
As a result, we see many successful single-person human pose estimation frameworks \cite{wei2016convolutional,newell2016stacked} with impressive results.
However, multi-person pose estimation remains a big challenge due to the lack of the ability for CNNs to reason about the relationship among detected keypoints.
In recent years, two major streams of approaches have been proposed to solve this challenging problem, \textit{i.e.}, top-down approaches \cite{he2017mask,papandreou2017towards, sun2018integral, chen2018cascaded, fang2017rmpe, huang2017coarse, xiao2018simple, sun2019deep} and bottom-up approaches \cite{cao2017realtime, newell2017associative, papandreou2018personlab, kreiss2019pifpaf, nie2018pose, nie2019singlestage}.

Top-down approaches first utilize a person detector to get multiple person bounding boxes.
A single-person pose estimator is then applied on each bounding box to detect the joints of the target person.
Top-down approaches are generally better at localizing keypoints as bounding boxes of different scales are normalized to the same scale for the second stage. In addition, the relationship between detected joints are implicitly represented by the bounding box since all joints within the bounding box are treated to be from the same person.
Despite the superior performance, top-down approaches are generally more computationally expensive because the single-person pose estimator has to be applied for each bounding box.
On the other hand, bottom-up approaches start with detecting keypoints from all person instances simultaneously in a single forward-pass, followed by a grouping stage where keypoints from the same person instance are grouped together. Bottom-up approaches are generally more efficient and less sensitive to the varying number of persons.

\begin{figure}[t]
    \centering
    \includegraphics[width=\columnwidth]{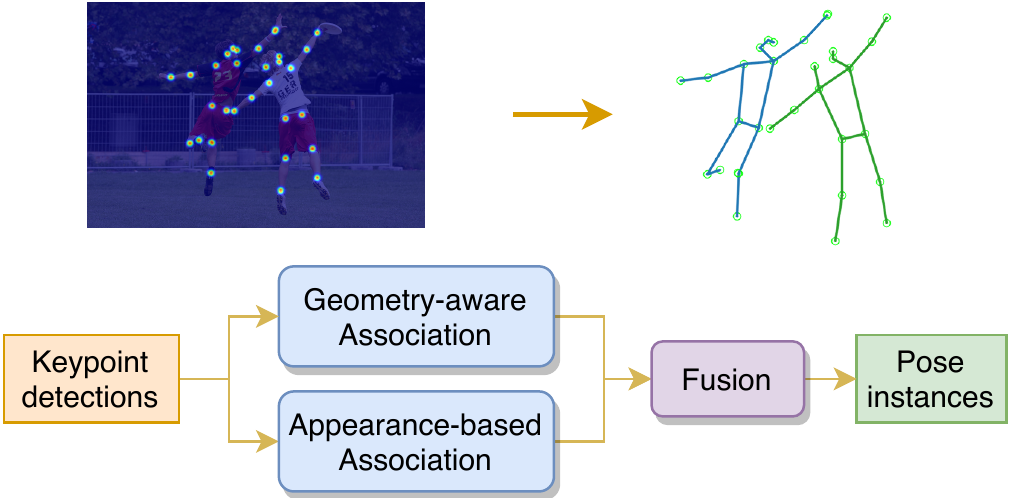}
    \caption{Our work tackles the grouping task in bottom-up multi-person pose estimation. We utilize a Geometry+Appearance association framework to group identity-free keypoint detections into different person instances.}
    \label{fig:teaser}
\end{figure}

Existing bottom-up approaches, \textit{e.g.} \cite{cao2017realtime,newell2017associative}, leverage on CNNs to learn specially-designed embeddings, which are used to compute the affinity between keypoint detections for grouping.
Despite the promising results from these works, several limitations still exist.
Firstly, the embedding (thus the affinity) is purely based on visual features, which is not robust to the diverse variations exhibited in real-world images.
Secondly, the learned embeddings are not good at handling instances with large varying scales, and the current approach to alleviate this problem to a limited extent is through multi-scale inference. Thirdly, 
the geometrical feasibility (anatomy) of human poses, \textit{e.g.}, possible angles, relative length between body parts, \textit{etc.}, is ignored because only visual features are utilized for grouping.

Inspired by the intuition that both spatial information and visual appearance are important 
for the challenging multi-person pose estimation problem, we propose a novel framework (Figure \ref{fig:teaser}) that fuses both geometry and appearance information to improve the grouping performance of bottom-up approaches. 
We formulate the grouping task as a graph partitioning problem and use a Graph Neural Network (GNN)-based framework to classify edges as either connected or disconnected. 
Specifically, we design a Geometry-aware Association Network which utilizes spatial information of the keypoints and learns local affinity from the global context. Unlike standard GNNs which rely on the graph topology to refine node representations, our framework learns the topology of the graph starting from an information-less fully-connected graph in an iterative manner. Moreover, we fuse the learned geometry-based affinity with the standard appearance-based affinity (similar as in \cite{cao2017realtime, newell2017associative}) to produce more robust association.
With the estimated graph output, standard spectral clustering can be applied to separate the clusters corresponding to different persons.
Our main contributions are:

\begin{itemize}
    \item We propose a GNN-based framework which leverages both spatial information and visual appearance to deal with the grouping task in bottom-up multi-person pose estimation. The proposed Geometry-aware Association Network learns topology context of the graph in contrast to standard GNNs which rely on the graph topology to propagate information inside the graph.
    \item We show performance gain over appearance-only grouping framework \cite{newell2017associative} consistently over multiple benchmark datasets and demonstrate the effectiveness in qualitative results. We also achieve competitive results on COCO multi-person task on par with the state-of-the-art.
\end{itemize}

\section{Related Work}

Extensive research has been done on the problem of multi-person pose estimation. Most approaches can be categorized into two branches: top-down approaches and bottom-up approaches.

\subsubsection{Top-down approaches}

Top-down approaches use bounding boxes detected by person detectors \cite{ren2015faster, redmon2016you, liu2016ssd, lin2017focal} to crop region of interest and detect keypoints for the single person inside the box.
Newell \textit{et al.} \cite{newell2016stacked} propose a deep keypoint detection framework with multiple stacked hourglass sub-networks showing great potential for human keypoints localization.
He \textit{et al.} \cite{he2017mask} extends the object detector framework \cite{ren2015faster} with an extra branch for keypoint detection.
More recent works including G-RMI \cite{papandreou2017towards}, RMPE \cite{fang2017rmpe}, CFN \cite{huang2017coarse}, CPN \cite{chen2018cascaded}, Simple baselines \cite{xiao2018simple}, and HRNet \cite{sun2019deep} propose various frameworks to increase the keypoint localization precision.

\subsubsection{Bottom-up approaches}

Bottom-up approaches detect all keypoints first followed by a grouping step to associate keypoints to person instances. 
Early approaches  \cite{insafutdinov2016deepercut, iqbal2016multi} exploit linear programming to solve for optimal association configuration.
With the development of Convolutional Neural Networks (CNNs), learned visual features have been incorporated into the grouping task.
OpenPose \cite{cao2017realtime} proposes Part Affinity Field which learns a 2D vector for each pixel encoding the evidence of body parts. These 2D vectors are sampled and integrated to compute the pairwise affinity for each pair of end keypoints of a body part.
Newell \textit{et al.} \cite{newell2017associative} propose Associative Embedding which learns a tag for each of the detected keypoints. A contrastive-like grouping loss is imposed to train the tag embedding so that distance between tag values can be used to easily distinguish keypoints from different person instances.
PersonLab \cite{papandreou2018personlab} extends previous works further to learn a geometric embedding where the offsets between keypoints are directly encoded in the feature maps and can be read out for grouping. Later works including Pose Partition Networks \cite{nie2018pose}, PifPaf \cite{kreiss2019pifpaf}, and Single-Stage Pose Machines \cite{nie2019singlestage} follow and further refine the idea of learning offset fields to help the task of grouping.

This work focuses on the grouping stage in bottom-up approaches. Unlike all previous grouping approaches that rely on the appearance-based affinity learned purely from visual features, our method attempts to incorporate geometric information into the grouping process which is able to understand global spatial configuration. Consequently, our approach is complementary to all appearance-based grouping approaches.

\subsubsection{Graph Neural Network}
Graph neural networks were first proposed to process data of graph structures in the form of recurrent neural networks \cite{gori2005new,scarselli2008graph,li2016gated}.
The idea of convolution on images is introduced to graph structured data \cite{duvenaud2015convolutional, kipf2017semi} where kernel parameters for aggregating features from neighboring nodes are shared across all locations in the graph.
Several approaches, \textit{e.g.}, GGT-NN \cite{johnson2017learning} and EGNN \cite{kim2019edge}, utilize specially-designed graph operations to tackle the problem of graph edge-labeling and show the strength of GNN on learning graph topology.
Compared to traditional GNNs where features of the nodes are used for graph message propagation, our Geometry-aware Association Network aggregates spatial context, \textit{i.e.}, embedding of neighboring edges, to form node representations and learns the graph topology from global spatial context.

\section{Approach}
\label{sec:approach}

\begin{figure*}[t]
    \centering
    \includegraphics[width=0.9\textwidth]{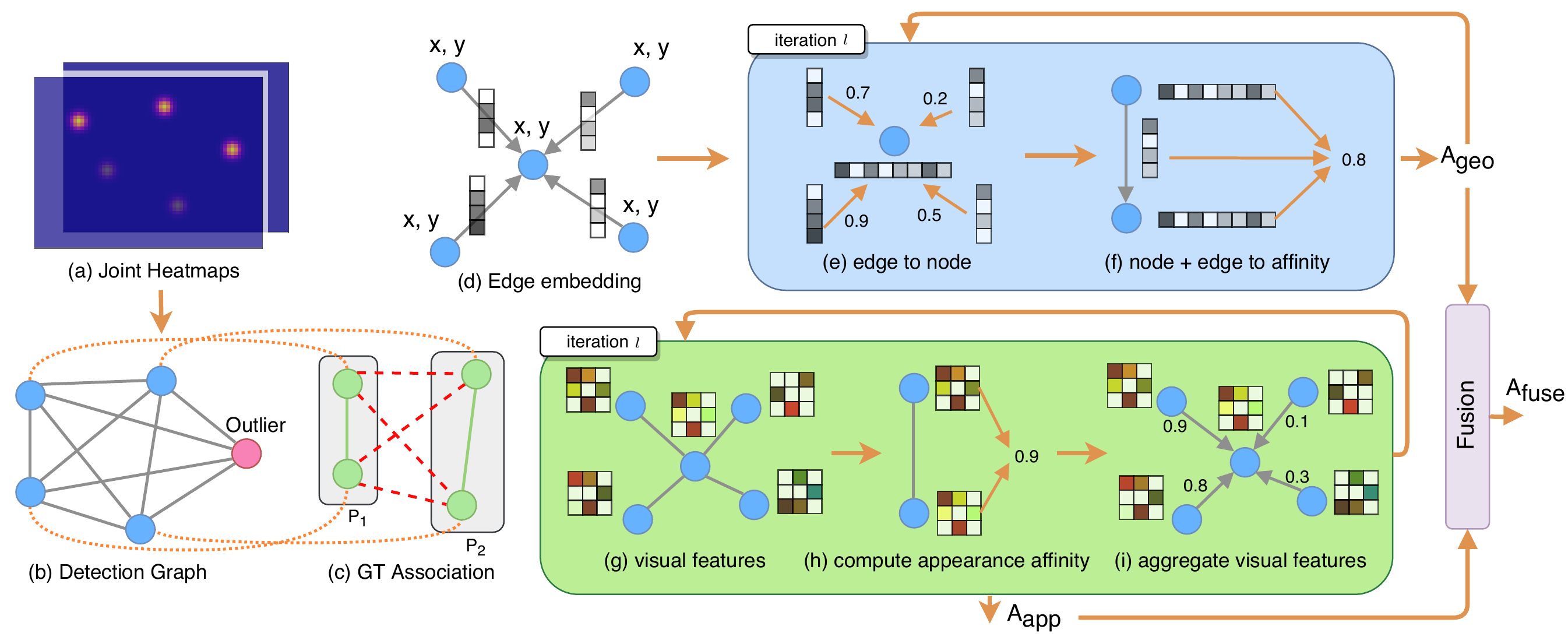}
    \caption{Our framework. (a)(b)(c) Graph is generated from detected keypoints. (d)(e)(f) Geometry-aware Association Network to learn association based on spatial information of the graph. Vertical bars represent edge embeddings. Horizontal bars represent node geometric features. (g)(h)(i) Appearance-based Association Network for visual feature clustering. Refer to text in Section \ref{sec:approach} for more details.}
    \label{fig:framework}
\end{figure*}

Given a 2D image with an unknown number of persons, the task of multi-person pose estimation is to identify the number of persons in the image and detect the pose of each person.
Following the bottom-up approaches, we utilize a human joint detector to detect all candidate joints and then group the joints into different persons. Formally, the detected body joints are denoted as $V=\{v_{1,1},..., v_{N_{1},1},v_{N_{1}+1,2},...,v_{N_{1}+N_{2},2},...,v_{N,J}\}$. Here, $J$ is the total number of joint types. $N_{j},~\forall j=1,2,...,J$, is the number of detected joints for each joint type $j$ and $N=\sum N_{j}$ is the total number of detected joints.
Given joint detections $V$, we generate a corresponding fully-connected graph and formulate the grouping task as a graph partitioning problem.
We design a GNN-based framework with a geometry-based branch and an appearance-based branch to jointly regress the probability of association for each pair of joints, represented by the output value of each edge in the graph. Spectral clustering is applied to partition the graph and we prune each cluster to get the final poses.
Figure \ref{fig:framework} shows an overview of our framework.

\subsection{Graph Generation}

We obtain the joint candidates by retrieving the local maxima on the output heatmaps of a joint detector (Figure \ref{fig:framework}(a)). For each detected joint $v_{m, j_m} \in V$ of type $j_m$, we generate a corresponding node in graph $G$ with following features: (1) $(x,y)$ coordinates of the joint in the image, both normalized to $[0,1]$, (2) joint type $j_m$, and (3) visual features extracted from a pretrained CNN. 
We form a undirected fully-connected graph $G(V, E)$ (Figure \ref{fig:framework}(b)) by connecting all pairs of nodes, \textit{i.e.}, $e_{m, n} \in E, \forall v_{m, j_m}, v_{n, j_n} \in V$.
We choose to generate edges between all pairs of joint types (\textit{e.g.}, head-to-hand, head-to-knee) because in the downstream clustering stage a fully-connected graph preserves the possibility to assign disjoint body parts into the same cluster. In contrast, graphs with only edges from the tree-structured topology of human body lead to separation of body parts from the same person if occlusion or missed detection of intermediate joints occurs.
In order to supervise learning of the model, we associate each of the detected joints to the ground truth joints set (Figure \ref{fig:framework}(c)). We assign a detection to a ground truth joint if the OKS (Object Keypoint Similarity, defined in COCO dataset \cite{lin2014microsoft}) is larger than a threshold. We allow multiple detections to be assigned to one ground truth joint as there may be duplicate detections due to the imperfection of the joint detector. We also allow detections to be assigned to none of the ground truth joints due to the missing ground truth annotation, which we denote as ``outliers". We design the following rules to label the edges of the graph:

\begin{enumerate}
    \item \label{itm: rule1} Label the edge as 1 if two detections are associated to ground truth joints of the same person.
    \item \label{itm: rule2} Label the edge as 0 if two detections are associated to ground truth joints of different persons.
    \item \label{itm: rule3} Mask out the edge during training if either of the two detections is an ``outlier". This avoids wrong penalization on unassigned detections caused by missing annotation from the dataset.
    \item \label{itm: rule4} We denote the set of labeled joint types as $L_{p}$ for each person $p$ in the ground truth, \textit{e.g.}, $L_{p}=\{\text{head}, \text{wrist}\}$. 
    Furthermore, we denote all detections assigned to $p$ as $I_{p}$ and all other detections of joint type within the set $L_{p}$ as $O_{p}$,  
    \textit{e.g.}, $I_{p}=\{v_{1,\text{head}}, v_{2,\text{wrist}}\}$ and 
    $O_{p}=\{v_{3,\text{head}}, v_{4,\text{head}}, v_{5,\text{wrist}}\}$. 
    Consequently, we label edge $e_{m,n}$ as 0,~$\forall v_{m, j_m} \in I_p, \forall v_{n, j_n} \in O_p$, since we are sure that nodes in $I_p$ and nodes in $O_p$ belong to different persons. 
\end{enumerate}
The rationale behind rule \ref{itm: rule4}: for example the joint ``ankle" is visible 
for person $p$,
but not labeled in ground truth, \textit{i.e.}, ankle $\not\in L_p$. Any ``ankle" detections not assigned to $p$ may still be the ankle of $p$. Therefore, we only impose supervision within the labeled set $L_p$.
Note that $O_p$ may contain ``outliers", thus rule \ref{itm: rule4} may override rule \ref{itm: rule3}.

\subsection{Geometry-aware Association Network}
Given a fully-connected graph with node features, the goal is to classify each of the edge as either 1 (connected) or 0 (not connected), so that nodes belonging to the same person are densely connected. Here, we propose a network that utilizes spatial information ($(x,y)$ coordinates of each joint) to learn feasible poses inside the graph. This is inspired by the fact that besides appearance similarity, human annotators also recognize human poses by identifying geometrical feasibility (such as relative displacement, relative lengths between different body parts, \textit{etc.}).
The network is illustrated in Figure \ref{fig:framework} (d),(e),(f).
We first encode the edges into a high-dimensional embedding (Figure \ref{fig:framework}(d)). For each pair of nodes $v_{m, j_{m}}$, $v_{n, j_{n}}$, the embedding is computed with a fully-connected network:
\begin{equation}
    \text{edge}_{m, n} = f_\text{edge}([x_m-x_n, y_m-y_n]; \Theta_{j_m, j_n}),
    \label{eqn:edge}
\end{equation}
where $\Theta_{j_m, j_n}$ are the parameters of the network.
Note that we use different parameters for different joint types combination $j_m$ and $j_n$ so that body parts information is also encoded in the embedding.

The edge embedding computed from equation \ref{eqn:edge} is fixed and shared across all iterations (Figure \ref{fig:framework}(d)).
For iteration $l$ of the network, the affinity matrix ${\mathbf{A}_\text{(geo)}}_{N\times N}^{(l-1)}$ from previous iteration is passed as input. Each joint will aggregate edges based on the normalized affinity score from $\mathbf{A}_\text{(geo)}$ (Figure \ref{fig:framework}(e)). Specifically, for a joint $v_{m, j_m}$, all incoming edges are aggregated and passed into a fully-connected network $f_\text{node}$ with parameters $\Theta_{j_m}$ to get the geometric features of the node:
\begin{equation}
    \begin{split}
        &\text{edge}, {\mathbf{A}_\text{(geo)}}^{(l-1)} \rightarrow \text{node}^{(l)}: \\
        &\text{node}_{m,j_m}^{(l)} = f_\text{node}\Big(\sum_{k} {{}\tilde{a}_\text{(geo)}}_{m, k}^{(l-1)}\cdot \text{edge}_{m, k}; \Theta_{j_m}\Big).
    \end{split}
\end{equation}
Here $\tilde{a}_\text{(geo)}$ are the normalized elements of $\mathbf{A}_\text{(geo)}$.
We $L1$-normalize within each joint type for each row of the affinity matrix since different joint types may have different number of detections.
The affinity score of each edge is then updated with another fully-connected network $f_\text{geo}$ with parameters $\Theta_{\text{geo}}$ (Figure \ref{fig:framework}(f)) by considering the features of two end nodes and the edge embedding:
\begin{equation}
    \begin{split}
        \text{edge}, &\text{node}^{(l)} \rightarrow {\mathbf{A}_\text{(geo)}}^{(l)}: {\mathbf{A}_\text{(geo)}}_{m,n}^{(l)} =\\
        &\sigma\Big(f_\text{geo}([\text{node}_{m,j_m}^{(l)}, \text{node}_{n,j_n}^{(l)},\text{edge}_{m,n}];\Theta_{\text{geo}})\Big),
    \end{split}
\end{equation}
and is passed on to the next iteration after a sigmoid activation $\sigma$. Note that since we do not have any prior knowledge of how to connect nodes in the graph, all elements in the input affinity matrix for the first iteration ${\mathbf{A}_\text{(geo)}}_{N\times N}^{(0)}$ are initialized to 1.

\subsection{Appearance-based Association Network and Fusion}

The Geometry-aware Association Network is already able to correctly predict most of the edges, as will be shown in Section \ref{sec:experiments}. However, in some difficult cases where multiple people are highly overlapped or cases where joint detections are not complete, predicting poses from spatial clue alone is not sufficient and we still need to resort to appearance to disambiguate such cases. We adopt the visual features (Figure \ref{fig:framework}(g)) extracted from the last layer of a CNN before output which is trained with the grouping loss as in Associative Embedding \cite{newell2017associative}. We use a standard Graph Neural Network to fine-tune the representation of the nodes by first computing the appearance affinity $\mathbf{A}_\text{(app)}$ (Figure \ref{fig:framework}(h)):
\begin{equation}
    \begin{split}
        &\text{node}^{(l-1)} \rightarrow {\mathbf{A}_\text{(app)}}^{(l)}: \\
        &{\mathbf{A}_\text{(app)}}_{m,n}^{(l)} = \sigma\Big[g_\text{app}\Big(\abs{\text{node}_{m,j_m}^{(l-1)} - \text{node}_{n,j_n}^{(l-1)}}; \Theta_{\text{app}}\Big)\Big],
    \end{split}
\end{equation}
and then aggregating neighboring nodes with the computed affinity (Figure \ref{fig:framework}(i)):
\begin{equation}
    \begin{split}
        &\text{node}^{(l-1)}, {\mathbf{A}_\text{(app)}}^{(l)} \rightarrow \text{node}^{(l)}: \text{node}_{m,j_m}^{(l)} =\\
        &g_\text{node}\Big([\sum_{k \ne m}{{}\tilde{a}_\text{(app)}}_{m,k}^{(l)}\cdot \text{node}_{k,j_k}^{(l-1)}, \text{node}_{m,j_m}^{(l-1)}]; \Theta_\text{node}\Big).
    \end{split}
\end{equation}
$\Theta_{\text{app}}$ and $\Theta_{\text{node}}$ are parameters of the two fully-connected networks $g_\text{app}$ and $g_\text{node}$, and $\tilde{a}_\text{(app)}$ are elements in row-wise $L1$-normalized affinity matrix $\mathbf{A}_\text{(app)}$.

We fuse $\mathbf{A}_\text{(geo)}$ and $\mathbf{A}_\text{(app)}$ after the last iteration of both branches through another fully-connected network $h_\text{fuse}$ to get the final predicted affinity matrix $\mathbf{A}_\text{(fuse)}$
\begin{equation}
    {\mathbf{A}_\text{(fuse)}}_{m,n}=h_\text{fuse}([{\mathbf{A}_\text{(geo)}}_{m,n}^{(L_\text{geo})}, {\mathbf{A}_\text{(app)}}_{m,n}^{(L_\text{app})}]; \Theta_\text{fuse}).
\end{equation}
We impose an edge-wise binary cross-entropy loss ($\text{BCE}$) on outputs of $\mathbf{A}_\text{(geo)}$ and $\mathbf{A}_\text{(app)}$ from each iteration as well as the final fused $\mathbf{A}_\text{(fuse)}$:
\begin{equation}
    \begin{split}
        \mathcal{L} = \sum_{e_{m,n} \in E} & \mathbf{M}_{m,n} \cdot 
        \Big(
        \text{BCE}({\mathbf{A}_\text{(fuse)}}_{m,n}, {\mathbf{A}_\text{(gt)}}_{m,n}) \\
        &+ \frac{1}{L_\text{geo}} \sum_{l=1}^{L_\text{geo}} \text{BCE}({\mathbf{A}_\text{(geo)}}_{m,n}^{(l)}, {\mathbf{A}_\text{(gt)}}_{m,n}) \\
        &+ \frac{1}{L_\text{app}}\sum_{l=1}^{L_\text{app}} \text{BCE}({\mathbf{A}_\text{(app)}}_{m,n}^{(l)}, {\mathbf{A}_\text{(gt)}}_{m,n})
        \Big).
    \end{split}
\end{equation}
$\mathbf{A}_\text{(gt)}$ is the ground truth affinity matrix with labels for each edge defined in rules \ref{itm: rule1}-\ref{itm: rule4}.
$\mathbf{M}$ is the mask matrix where $\mathbf{M}_{m,n} = 0$ if $e_{m,n}$ satisfies rule \ref{itm: rule3} but not rule \ref{itm: rule4}, or $\mathbf{M}_{m,n} = 1$ otherwise.
$L_\text{geo}$ and $L_\text{app}$ are the number of iterations for Geometry-aware Association Network and Appearance-based Association Network respectively.

\subsection{Grouping}

After the affinity matrix is obtained, we extract poses by partitioning the graph into highly intra-connected clusters with spectral clustering. We first binarize all the edges in the affinity matrix with a threshold 0.5. We then compute the Laplacian matrix and conduct the eigendecomposition thereon. The number of clusters can be effectively estimated by counting the number of close-to-zero eigenvalues. K-means is used to get the clusters from the eigenvectors. Within each cluster, we select a subset of nodes as a candidate pose so that: (1) there is at most one node selected from each joint type, and (2) no selected node in the subset can be replaced by node of the same joint type with a higher average affinity to the selected subset.
This guarantees that even though joints from more than one person are included in a cluster, we can still accurately select the pose with compactly associated joints.
We repeat this process for nodes that are not selected to generate more pose candidates until all nodes are processed.

\section{Experiments}
\label{sec:experiments}

\subsection{Datasets and implementation details}
We conduct experiments on COCO \cite{lin2014microsoft} and MPII \cite{andriluka14cvpr} datasets for the multi-person pose estimation task. COCO dataset contains over 150k images and each person instance is annotated with 17 keypoints. It is divided into \{\textit{train}, \textit{val}, \textit{test-dev}\} sets, where each contains 64k, 2.6k, and 20k images with person keypoint annotations. We report the results in the standard average precision and average recall metrics which are computed based on Object Keypoint Similarity (OKS). MPII dataset is another in-the-wild 2D human pose dataset with large appearance variation. We directly apply our model trained with COCO dataset on a 350-images validation set to evaluate the generalizability of our model.

We follow \cite{newell2017associative} and use a Stacked Hourglass Network to generate joint detections. We apply Non-Maximum-Suppression (NMS) on the predicted heatmaps and only keep joints with confidence $\geq 0.1$. Detected joints with OKS $\geq 0.5$ to any ground-truth joint are associated to ground-truth person instances. For geometry-aware network, we use 256-channel vectors for both the edge embedding and the nodes' aggregated features. For appearance-based network, we extract a 3x3 patch from the last layer of the last hourglass surrounding each detection and map it to a 256-dimension hidden space. To extract features for different joint types, the mapping function uses different set of parameters for each of the $J$ joint types. We set the number of iterations to 3 for the geometry-aware network and 2 for the appearance-based network, since we empirically found that adding more iterations do not give further performance gain.
Weights are not shared across iterations.
We use two fully-connected layers with 256 hidden units, each followed by ReLU activation, for all sub-networks $f_\text{edge}$, $f_\text{node}$, $f_\text{geo}$, $g_\text{app}$, $g_\text{node}$, except for the last layer of $f_\text{geo}$ and $g_\text{app}$ which outputs a scalar affinity value with sigmoid activation.
The fusion network $h_\text{fuse}$ takes both $\textbf{A}_\text{(geo)}$ and $\textbf{A}_\text{(app)}$ before sigmoid from their respective last iteration as input and consists of fully-connected layers with 2-16-64-64-16-1 hidden units followed by a sigmoid layer.
We do not form mini-batches but set batch size to 1 because the graph size varies among different images. We follow previous works \cite{cao2017realtime,newell2017associative,nie2019singlestage} to refine estimation results with a single-person pose estimation model trained on COCO. We only update for keypoints if the original and refined location is close (within 5 pixel distance).
In evaluation, the inference of our GNN model takes $\sim$ 7ms per image, and spectral clustering takes $\sim$ 38ms per image.

\begin{figure}[t]
    \centering
    \includegraphics[width=\columnwidth]{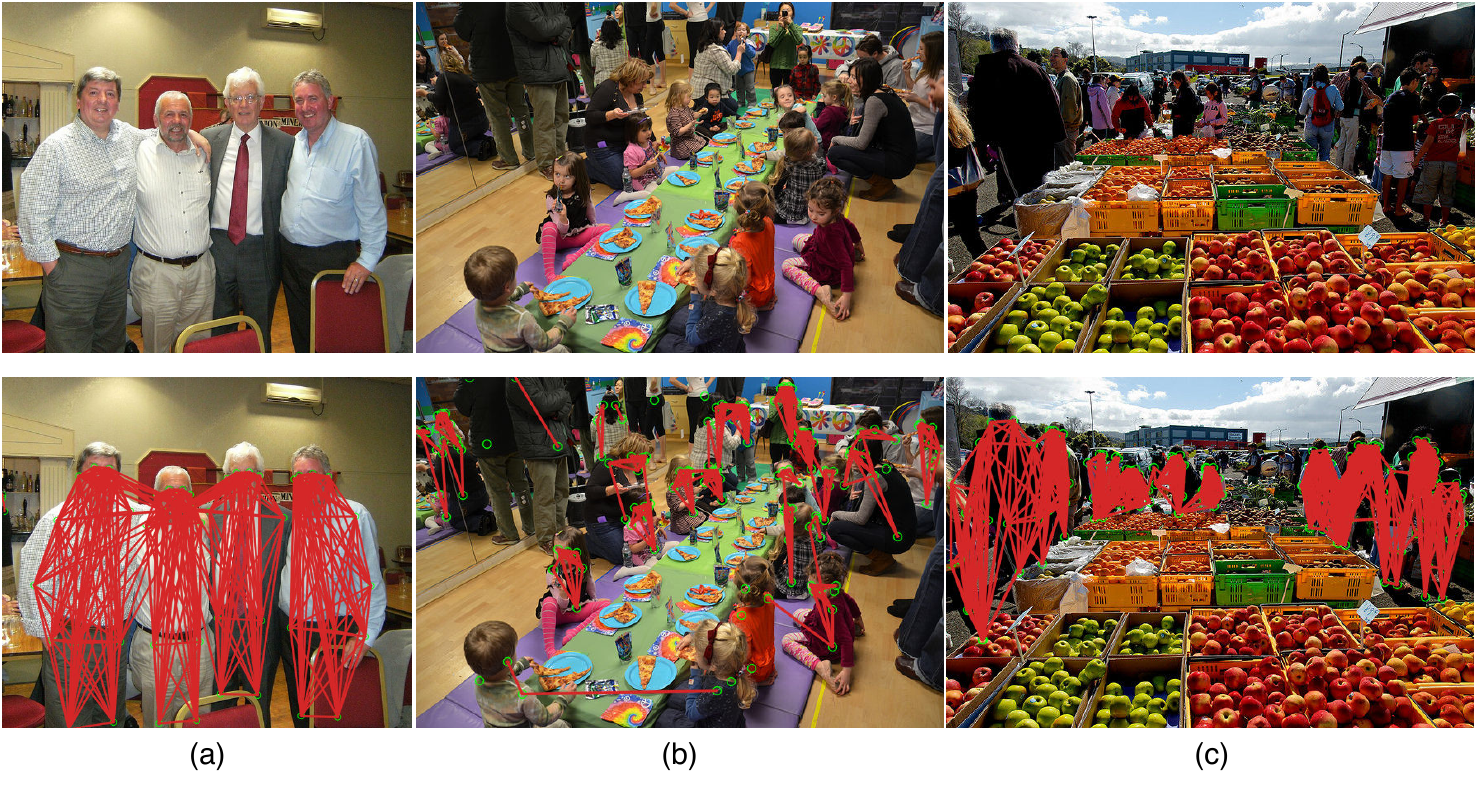}
    \caption{Association results considering only spatial information. First row are original images. Second row are the graph outputs from Geometry-aware Association Network. Red lines indicate edges predicted to be connected.}
    \label{fig:geo-only}
\end{figure}

\subsection{Geometry-aware Association}

We first examine the effectiveness of the Geometry-aware Association Network by observing the affinity matrix estimated by this network, \textit{i.e.}, $\mathbf{A}_\text{(geo)}$. Using only spatial information ($x, y$ coordinates), the Geometry-aware Association Network achieves above 97\% accuracy when evaluating the proportion of edges in the graph that are correctly classified. As is shown in Figure \ref{fig:geo-only}, the network successfully distinguishes between different person instances in most cases. Even in complicated cases such as in Figure \ref{fig:geo-only}(b), we can still see most cluster patterns identified and separated in a meaningful way. Meanwhile, we find that association errors tend to happen when detected joints are not complete for a particular person. As an example, half of the body for the leftmost person in Figure \ref{fig:geo-only}(c) is missing in the graph and it leads to confusion for the network to correctly partition the graph with spatial information alone.
Results above show that it is effective to consider global context with spatial information, while we still need extra information such as visual appearance to help to disambiguate local associations.

\begin{table}[t]
    \centering
    \caption{Comparison with appearance-based grouping approach \cite{newell2017associative}. $+$ means using refinement.}
    \label{tab:compare-with-ae}
    \begin{adjustbox}{max width=\columnwidth}
        \begin{tabular}{c|c|c|c|c|c|c|c}
            \hline
            \noalign{\smallskip}
            Method & Split & AP & $\text{AP}^{50}$ & $\text{AP}^{75}$ & $\text{AP}^{M}$ & $\text{AP}^{L}$ & AR \\
            \noalign{\smallskip}
            \hline
            \noalign{\smallskip}
            AE \cite{newell2017associative} & \textit{val} & 67.6 & 86.6 & 73.4 & 62.2 & 76.0 & 71.8 \\
            Ours & \textit{val} & \textbf{69.8} & \textbf{87.8} & \textbf{76.0} & \textbf{64.4} & \textbf{78.0} & \textbf{73.4} \\
            \noalign{\smallskip}
            \hline
            \noalign{\smallskip}
            AE \cite{newell2017associative} & \textit{test-dev} & 63.0 & 85.7 & 68.9 & 58.0 & 70.4 & - \\
            Ours & \textit{test-dev} & \textbf{64.7} & \textbf{86.4} & \textbf{71.0} & \textbf{59.2} & \textbf{73.1} & \textbf{69.8} \\
            \noalign{\smallskip}
            \hline
            \noalign{\smallskip}
            AE \cite{newell2017associative} $+$ & \textit{test-dev} & 65.5 & 86.8 & 72.3 & 60.6 & 72.6 & 70.2 \\
            Ours $+$ & \textit{test-dev} & \textbf{68.2} & \textbf{86.9} & \textbf{75.4} & \textbf{64.8} & \textbf{74.2} & \textbf{73.3} \\
            \noalign{\smallskip}
            \hline
        \end{tabular}
    \end{adjustbox}
\end{table}

\begin{table*}[t]
    \centering
    \caption{Comparison with bottom-up approaches on COCO \textit{test-dev} split. $+$ means using refinement with single-person pose estimator. Best results are bold-faced. Second-best are underlined.}
    \label{tab:benchmark}
    \begin{adjustbox}{max width=0.9\textwidth}
        \begin{tabular}{l|l|c|c|c|c|c|c|c|c|c|c}
            \hline
            \noalign{\smallskip}
            Method & Detector & AP & $\text{AP}^{50}$ & $\text{AP}^{75}$ & $\text{AP}^{M}$ & $\text{AP}^{L}$ & AR & $\text{AR}^{50}$ & $\text{AR}^{75}$ & $\text{AR}^{M}$ & $\text{AR}^{L}$\\
            \noalign{\smallskip}
            \hline
            \noalign{\smallskip}
            OpenPose \cite{cao2017realtime} $+$ & - & 61.8 & 84.9 & 67.5 & 57.1 & 68.2 & 66.5 & 87.2 & 71.8 & 60.6 & 74.6 \\
            AE \cite{newell2017associative} $+$ & Hourglass & 65.5 & 86.8 & 72.3 & 60.6 & 72.6 & 70.2 & 89.5 & 76.0 & 64.6 & 78.1 \\
            PifPaf \cite{kreiss2019pifpaf} & ResNet-152 & 66.7 & - & - & 62.4 & 72.9 & - & - & - & - & - \\
            Single-Stage \cite{nie2019singlestage} $+$ & Hourglass & 66.9 & \underline{88.5} & 72.9 & 62.6 & 73.1 & - & - & - & - & - \\
            PersonLab \cite{papandreou2018personlab} & ResNet-152 & \textbf{68.7} & \textbf{89.0} & \textbf{75.4} & \underline{64.1} & \textbf{75.5} & \textbf{75.4} & \textbf{92.7} & \textbf{81.2} & \textbf{69.7} & \textbf{83.0} \\
            \noalign{\smallskip}
            \hline
            \noalign{\smallskip}
            Ours $+$ & Hourglass & \underline{68.2} & 86.9 & \textbf{75.4} & \textbf{64.8} & \underline{74.2} & \underline{73.3} & \underline{89.8} & \underline{79.7} & \underline{69.0} & \underline{79.3} \\
            \noalign{\smallskip}
            \hline
        \end{tabular}
    \end{adjustbox}
\end{table*}

\subsection{Quantitative results on COCO dataset}

We compare with Associative Embedding \cite{newell2017associative} to show the advantage of incorporating spatial information for better grouping.
We choose \cite{newell2017associative} as our comparison baseline because: 
(1) it is the most representative grouping approach that utilizes only appearance features;
and (2) it is the best state-of-the-art approach with publicly available
source code for the joint detector
since we have to keep the joint detections the same for a fair comparison on the grouping part.
Table \ref{tab:compare-with-ae} shows the results on both \textit{val} and \textit{test-dev} split. While using the same detections as in \cite{newell2017associative}, our geometry+appearance grouping approach outperforms the appearance-only approach on both splits by a competitive margin (2.2\% AP on \textit{val} and 1.7\% AP on \textit{test-dev}).

We report the performance with the full list of bottom-up approaches in Table \ref{tab:benchmark}. Our framework achieves overall 68.2\% AP which is on par with the state-of-the-art \cite{papandreou2018personlab}. It is worth pointing out that although we use the same hourglass-based backbone from \cite{newell2017associative,nie2019singlestage} as keypoint detector, our grouping algorithm shows comparable results to the state-of-the-art \cite{papandreou2018personlab} which uses a multi-stage refinement scheme to localize keypoints with higher localization precision.
As localization error is a major source of error pointed out by \cite{ruggero2017benchmarking}, our grouping framework has the potential to gain further performance boost with a better backbone keypoint detector.

\begin{table}[t]
    \centering
    \caption{Results on different keypoint detections on COCO \textit{val} split. \cite{sun2019deep} is a top-down approach where bounding boxes implicitly associate keypoints.}
    \label{tab:better-detections}
    \begin{adjustbox}{max width=\columnwidth}
        \begin{tabular}{c|c|c|c|c}
            \hline
            \noalign{\smallskip}
            Method & Detections & Grouping & AP & AR \\
            \noalign{\smallskip}
            \hline
            \noalign{\smallskip}
            HRNet \cite{sun2019deep} & HRNet & - & 76.6 & 81.2 \\
            \noalign{\smallskip}
            \hline
            \noalign{\smallskip}
            AE \cite{newell2017associative} & HRNet & AE & 65.6 & 71.7 \\
            Ours & HRNet & GNN & \textbf{75.7} & \textbf{78.9} \\
            \noalign{\smallskip}
            \hline
            \noalign{\smallskip}
            AE \cite{newell2017associative} & GT & AE & 90.9 & 93.2 \\
            Ours & GT & GNN & \textbf{95.3} & \textbf{96.4} \\
            \noalign{\smallskip}
            \hline
        \end{tabular}
    \end{adjustbox}
\end{table}

\begin{table}[t]
    \centering
    \caption{Results on MPII validation set (AP).}
    \label{tab:mpii}
    \begin{adjustbox}{max width=\columnwidth}
        \begin{tabular}{c|cccccc|c}
            \hline
            \noalign{\smallskip}
            Method & Shoulder & Elbow & Wrist & Hip & Knee & Ankle & Total \\
            \noalign{\smallskip}
            \hline
            \noalign{\smallskip}
            AE \cite{newell2017associative} & 80.0 & 71.0 & \textbf{62.4} & 70.0 & 63.0 & \textbf{53.4} & 66.6\\
            Ours & \textbf{85.7} & \textbf{71.1} & 60.7 & \textbf{73.1} & \textbf{64.6} & 53.2 & \textbf{68.1}\\
            \noalign{\smallskip}
            \hline
        \end{tabular}
    \end{adjustbox}
\end{table}

\subsection{Grouping performance with better keypoint detections}

We perform an additional experiment to apply our framework on other sets of detections with different levels of keypoint localization accuracy. We conduct the experiments on: (1) the detected joints from the current state-of-the-art top-down approach by Sun \textit{et al.} \cite{sun2019deep} which utilizes higher-resolution feature representations for finer keypoint localization, and (2) the ground truth annotations of the keypoint locations. Specifically, for the former setting, we follow the standard top-down approach pipeline to detect the keypoints for each of the person bounding boxes. Instead of assembling all the keypoints within each bounding box into a human instance, we drop the bounding box information and collect all the keypoints from all bounding boxes into a big keypoint candidates pool. We do a non-maximum-suppression for the keypoint candidates to remove redundant keypoint detections.
For the ground truth annotations, we simply treat the identity of all the keypoints as unknown and only use the location of each of the keypoints.
Given the detections, we create a corresponding detection graph and perform the association with our framework.
It is worth to note that our GNN-based model does not require re-training when applied to different sources of detections.
Results are shown in Table \ref{tab:better-detections}.
We see that when using detections from the top-down single-person detector, our graph-based framework achieves performance almost on par with the original top-down approach \cite{sun2019deep} which has inherent grouping information from the bounding box. By using the ground truth keypoints, we can further achieve 95.3\% AP on the \textit{val} set.
Both results show that despite being a bottom-up approach, our grouping framework has the potential of achieving good performances given better keypoint detections. 
Additionally, our framework significantly outperforms the appearance-only grouping approach AE \cite{newell2017associative} on both sets of keypoints.

\subsection{Results on MPII dataset}
To evaluate the generalizability of our framework, we apply our GNN-based model trained with COCO dataset directly on MPII validation set without further training;
and compare with AE \cite{newell2017associative} using the same joint detector pretrained on COCO.
Average precision for all joints except ``Head" are reported due to the joint set difference between the two datasets.
Results are shown in Table \ref{tab:mpii} where our grouping framework outperforms \cite{newell2017associative} by 1.5\% in overall AP.

\begin{figure}[t]
    \centering
    \includegraphics[width=0.9\columnwidth]{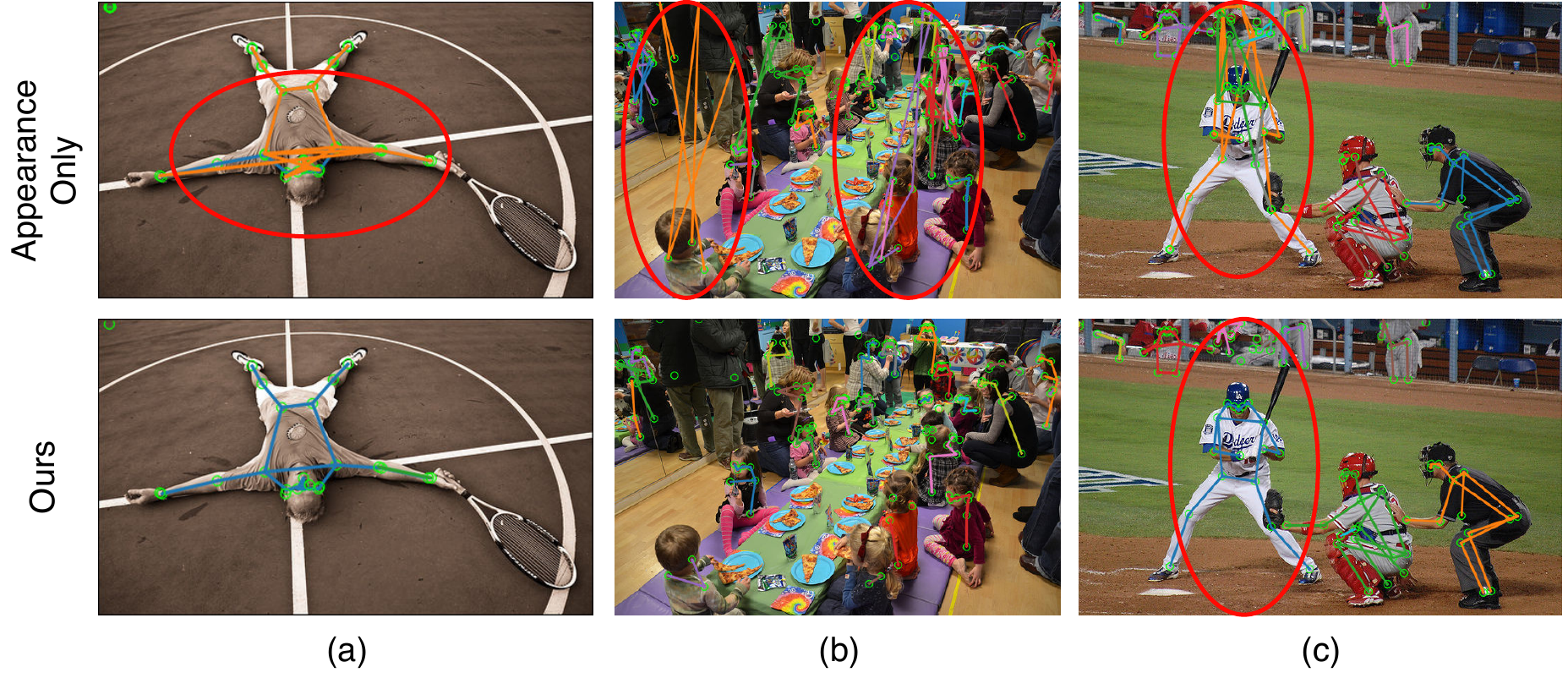}
    \caption{Qualitative results on COCO \textit{val} split compared with the results from \cite{newell2017associative}. Poses of different persons are shown in different colors}
    \label{fig:qualitative}
\end{figure}

\subsection{Qualitative results}
\label{sec:qualitative}

We show qualitative results on COCO \textit{val} set in Figure \ref{fig:qualitative}.
Ignoring spatial information and relying on visual features alone leads to invalid poses, while our approach effectively corrects the grouping in most cases.

Our geometry-aware framework also exhibits association errors, most of which are caused by incomplete keypoint detections. Our result in Figure \ref{fig:qualitative}(b) is an example where some of the edges are missing due to a lot of partially visible poses in the image. Robustness to partial detections is a key direction for future improvement of our framework.

\section{Conclusion}

In this work, we proposed a novel graph-based framework to tackle the grouping task for bottom-up multi-person pose estimation. We formulated the grouping task as a graph partitioning problem and designed a Geometry-aware Association Network to learn the spatial configuration of human poses in the graph. We fused geometry-based affinity with appearance-based affinity and applied spectral clustering to obtain poses for different person instances. We demonstrated the effectiveness of our framework with improved results over the appearance-only grouping approach.

% \addtolength{\textheight}{-12cm}   % This command serves to balance the column lengths
%                                   % on the last page of the document manually. It shortens
%                                   % the textheight of the last page by a suitable amount.
%                                   % This command does not take effect until the next page
%                                   % so it should come on the page before the last. Make
%                                   % sure that you do not shorten the textheight too much.

%%%%%%%%%%%%%%%%%%%%%%%%%%%%%%%%%%%%%%%%%%%%%%%%%%%%%%%%%%%%%%%%%%%%%%%%%%%%%%%%

%%%%%%%%%%%%%%%%%%%%%%%%%%%%%%%%%%%%%%%%%%%%%%%%%%%%%%%%%%%%%%%%%%%%%%%%%%%%%%%%

%%%%%%%%%%%%%%%%%%%%%%%%%%%%%%%%%%%%%%%%%%%%%%%%%%%%%%%%%%%%%%%%%%%%%%%%%%%%%%%%
% \section*{APPENDIX}

% Appendixes should appear before the acknowledgment.

% \section*{ACKNOWLEDGMENT}

% The preferred spelling of the word ÒacknowledgmentÓ in America is without an ÒeÓ after the ÒgÓ. Avoid the stilted expression, ÒOne of us (R. B. G.) thanks . . .Ó  Instead, try ÒR. B. G. thanksÓ. Put sponsor acknowledgments in the unnumbered footnote on the first page.

%%%%%%%%%%%%%%%%%%%%%%%%%%%%%%%%%%%%%%%%%%%%%%%%%%%%%%%%%%%%%%%%%%%%%%%%%%%%%%%%

% References are important to the reader; therefore, each citation must be complete and correct. If at all possible, references should be commonly available publications.

\bibliographystyle{IEEETran}
\bibliography{root}

\begin{thebibliography}{10}
\providecommand{\url}[1]{#1}
\csname url@rmstyle\endcsname
\providecommand{\newblock}{\relax}
\providecommand{\bibinfo}[2]{#2}
\providecommand\BIBentrySTDinterwordspacing{\spaceskip=0pt\relax}
\providecommand\BIBentryALTinterwordstretchfactor{4}
\providecommand\BIBentryALTinterwordspacing{\spaceskip=\fontdimen2\font plus
\BIBentryALTinterwordstretchfactor\fontdimen3\font minus
  \fontdimen4\font\relax}
\providecommand\BIBforeignlanguage[2]{{%
\expandafter\ifx\csname l@#1\endcsname\relax
\typeout{** WARNING: IEEEtran.bst: No hyphenation pattern has been}%
\typeout{** loaded for the language `#1'. Using the pattern for}%
\typeout{** the default language instead.}%
\else
\language=\csname l@#1\endcsname
\fi
#2}}

\bibitem{wei2016convolutional}
S.-E. Wei, V.~Ramakrishna, T.~Kanade, and Y.~Sheikh, ``Convolutional pose
  machines,'' in \emph{IEEE Conference on Computer Vision and Pattern
  Recognition}, 2016.

\bibitem{newell2016stacked}
A.~Newell, K.~Yang, and J.~Deng, ``Stacked hourglass networks for human pose
  estimation,'' in \emph{European Conference on Computer Vision}, 2016.

\bibitem{he2017mask}
K.~He, G.~Gkioxari, P.~Doll{\'a}r, and R.~Girshick, ``Mask r-cnn,'' in
  \emph{International Conference on Computer Vision}, 2017.

\bibitem{papandreou2017towards}
G.~Papandreou, T.~Zhu, N.~Kanazawa, A.~Toshev, J.~Tompson, C.~Bregler, and
  K.~Murphy, ``Towards accurate multi-person pose estimation in the wild,'' in
  \emph{IEEE Conference on Computer Vision and Pattern Recognition}, 2017.

\bibitem{sun2018integral}
X.~Sun, B.~Xiao, F.~Wei, S.~Liang, and Y.~Wei, ``Integral human pose
  regression,'' in \emph{European Conference on Computer Vision}, 2018.

\bibitem{chen2018cascaded}
Y.~Chen, Z.~Wang, Y.~Peng, Z.~Zhang, G.~Yu, and J.~Sun, ``Cascaded pyramid
  network for multi-person pose estimation,'' in \emph{IEEE Conference on
  Computer Vision and Pattern Recognition}, 2018.

\bibitem{fang2017rmpe}
H.-S. Fang, S.~Xie, Y.-W. Tai, and C.~Lu, ``Rmpe: Regional multi-person pose
  estimation,'' in \emph{International Conference on Computer Vision}, 2017.

\bibitem{huang2017coarse}
S.~Huang, M.~Gong, and D.~Tao, ``A coarse-fine network for keypoint
  localization,'' in \emph{International Conference on Computer Vision}, 2017.

\bibitem{xiao2018simple}
B.~Xiao, H.~Wu, and Y.~Wei, ``Simple baselines for human pose estimation and
  tracking,'' in \emph{European Conference on Computer Vision}, 2018.

\bibitem{sun2019deep}
K.~Sun, B.~Xiao, D.~Liu, and J.~Wang, ``Deep high-resolution representation
  learning for human pose estimation,'' in \emph{IEEE Conference on Computer
  Vision and Pattern Recognition}, 2019.

\bibitem{cao2017realtime}
Z.~Cao, T.~Simon, S.-E. Wei, and Y.~Sheikh, ``Realtime multi-person 2d pose
  estimation using part affinity fields,'' in \emph{IEEE Conference on Computer
  Vision and Pattern Recognition}, 2017.

\bibitem{newell2017associative}
A.~Newell, Z.~Huang, and J.~Deng, ``Associative embedding: End-to-end learning
  for joint detection and grouping,'' in \emph{Conference on Neural Information
  Processing Systems}, 2017.

\bibitem{papandreou2018personlab}
G.~Papandreou, T.~Zhu, L.-C. Chen, S.~Gidaris, J.~Tompson, and K.~Murphy,
  ``Personlab: Person pose estimation and instance segmentation with a
  bottom-up, part-based, geometric embedding model,'' in \emph{European
  Conference on Computer Vision}, 2018.

\bibitem{kreiss2019pifpaf}
S.~Kreiss, L.~Bertoni, and A.~Alahi, ``Pifpaf: Composite fields for human pose
  estimation,'' in \emph{IEEE Conference on Computer Vision and Pattern
  Recognition}, 2019.

\bibitem{nie2018pose}
X.~Nie, J.~Feng, J.~Xing, and S.~Yan, ``Pose partition networks for
  multi-person pose estimation,'' in \emph{European Conference on Computer
  Vision}, 2018.

\bibitem{nie2019singlestage}
X.~Nie, J.~Zhang, S.~Yan, and J.~Feng, ``Single-stage multi-person pose
  machines,'' in \emph{International Conference on Computer Vision}, 2019.

\bibitem{ren2015faster}
S.~Ren, K.~He, R.~Girshick, and J.~Sun, ``Faster r-cnn: Towards real-time
  object detection with region proposal networks,'' in \emph{Conference on
  Neural Information Processing Systems}, 2015.

\bibitem{redmon2016you}
J.~Redmon, S.~Divvala, R.~Girshick, and A.~Farhadi, ``You only look once:
  Unified, real-time object detection,'' in \emph{IEEE Conference on Computer
  Vision and Pattern Recognition}, 2016.

\bibitem{liu2016ssd}
W.~Liu, D.~Anguelov, D.~Erhan, C.~Szegedy, S.~Reed, C.-Y. Fu, and A.~C. Berg,
  ``Ssd: Single shot multibox detector,'' in \emph{European Conference on
  Computer Vision}, 2016.

\bibitem{lin2017focal}
T.-Y. Lin, P.~Goyal, R.~Girshick, K.~He, and P.~Doll{\'a}r, ``Focal loss for
  dense object detection,'' in \emph{International Conference on Computer
  Vision}, 2017.

\bibitem{insafutdinov2016deepercut}
E.~Insafutdinov, L.~Pishchulin, B.~Andres, M.~Andriluka, and B.~Schiele,
  ``Deepercut: A deeper, stronger, and faster multi-person pose estimation
  model,'' in \emph{European Conference on Computer Vision}, 2016.

\bibitem{iqbal2016multi}
U.~Iqbal and J.~Gall, ``Multi-person pose estimation with local joint-to-person
  associations,'' in \emph{European Conference on Computer Vision}, 2016.

\bibitem{gori2005new}
M.~Gori, G.~Monfardini, and F.~Scarselli, ``A new model for learning in graph
  domains,'' in \emph{International Joint Conference on Neural Networks}, 2005.

\bibitem{scarselli2008graph}
F.~Scarselli, M.~Gori, A.~C. Tsoi, M.~Hagenbuchner, and G.~Monfardini, ``The
  graph neural network model,'' \emph{IEEE Transactions on Neural Networks},
  vol.~20, no.~1, pp. 61--80, 2008.

\bibitem{li2016gated}
Y.~Li, D.~Tarlow, M.~Brockschmidt, and R.~Zemel, ``Gated graph sequence neural
  networks,'' in \emph{International Conference on Learning Representations},
  2016.

\bibitem{duvenaud2015convolutional}
D.~K. Duvenaud, D.~Maclaurin, J.~Iparraguirre, R.~Bombarell, T.~Hirzel,
  A.~Aspuru-Guzik, and R.~P. Adams, ``Convolutional networks on graphs for
  learning molecular fingerprints,'' in \emph{Conference on Neural Information
  Processing Systems}, 2015.

\bibitem{kipf2017semi}
T.~N. Kipf and M.~Welling, ``Semi-supervised classification with graph
  convolutional networks,'' in \emph{International Conference on Learning
  Representations}, 2017.

\bibitem{johnson2017learning}
D.~D. Johnson, ``Learning graphical state transitions,'' in \emph{International
  Conference on Learning Representations}, 2017.

\bibitem{kim2019edge}
J.~Kim, T.~Kim, S.~Kim, and C.~D. Yoo, ``Edge-labeling graph neural network for
  few-shot learning,'' in \emph{IEEE Conference on Computer Vision and Pattern
  Recognition}, 2019.

\bibitem{lin2014microsoft}
T.-Y. Lin, M.~Maire, S.~Belongie, J.~Hays, P.~Perona, D.~Ramanan,
  P.~Doll{\'a}r, and C.~L. Zitnick, ``Microsoft coco: Common objects in
  context,'' in \emph{European Conference on Computer Vision}, 2014.

\bibitem{andriluka14cvpr}
M.~Andriluka, L.~Pishchulin, P.~Gehler, and B.~Schiele, ``2d human pose
  estimation: New benchmark and state of the art analysis,'' in \emph{IEEE
  Conference on Computer Vision and Pattern Recognition}, June 2014.

\bibitem{ruggero2017benchmarking}
M.~Ruggero~Ronchi and P.~Perona, ``Benchmarking and error diagnosis in
  multi-instance pose estimation,'' in \emph{International Conference on
  Computer Vision}, 2017.

\end{thebibliography}

\end{document}